\documentclass[letterpaper, 10 pt, conference]{Templates/ieeeconf}
\pdfoutput=1
\IEEEoverridecommandlockouts
\usepackage[utf8]{inputenc}
\usepackage{hyperref}
\usepackage{xcolor}
\usepackage{graphicx}
\usepackage{amsmath,amsfonts}
\usepackage{multirow}
\usepackage{subcaption}

\usepackage{comment}
\usepackage{pbox}

\usepackage{setspace}
\let\labelindent\relax
\usepackage{enumitem}
\usepackage{algorithm,algpseudocode}
\usepackage{booktabs}
\usepackage{color}
\usepackage{cite}
\usepackage{amssymb}
\usepackage{array}
\usepackage{siunitx}
\usepackage{xparse,mathtools}

\usepackage{acronym}
\acrodef{WVU}{West Virginia University}
\acrodef{ROS}{Robot Operating System}
\acrodef{IMU}{Inertial Measurement Unit}
\acrodef{SLAM}{Simultaneous Localization and Mapping}
\acrodef{TSP}{Traveling Salesmen Problem}
\acrodef{FSM}{finite-state machine}
\acrodef{FCL}{Flexible Collision Library}
\acrodef{ISS}{International Space Station}

\DeclareMathOperator*{\argmax}{argmax}
\DeclareMathOperator*{\argmin}{argmin}

\title{\LARGE \bf
Flower Interaction Subsystem for a Precision Pollination Robot
}

\author{Jared Strader$^{1}$, Jennifer Nguyen$^{1}$, Christopher Tatsch$^{1}$, Yixin Du$^{2}$, Kyle Lassak$^{1}$, Benjamin Buzzo$^{1}$,\\ Ryan Watson$^{1}$, Henry Cerbone$^{1}$, Nicholas Ohi$^{1}$, Chizhao Yang$^{1}$, and Yu Gu$^{1}$
\thanks{$^{1}$Department of Mechanical and Aerospace Engineering, West Virginia University, Morgantown, WV}%
\thanks{$^{2}$Lane Department of Computer Science and Electrical Engineering, West Virginia University, Morgantown, WV}
\thanks{$^{*}$ This research was supported in part by the United States Department of Agriculture (USDA) National Institute of Food and Agriculture (NIFA) Project: 2017-67022-25926 and the Arlen G. \& Louise Stone Swiger Fellowship.}
}

\begin{document}

\renewcommand{\figureautorefname}{Fig.}
\renewcommand{\tableautorefname}{Table}
\renewcommand{\sectionautorefname}{Section}
\renewcommand{\subsectionautorefname}{Section}

\begin{onecolumn}
\Huge{\textbf{IEEE Copyright Notice}}

\vspace{2.5em}

\noindent\large{\textcopyright 2019 IEEE. Personal use of this material is permitted. Permission from IEEE must be obtained for all other uses, in any current or future media, including reprinting/republishing this material for advertising or promotional purposes, creating new collective works, for resale or redistribution to servers or lists, or reuse of any copyrighted component of this work in other works.}

\vspace{5em}

\noindent\Large{Accepted to be published in: Proceedings of the 2019 IEEE/RSJ International Conference on Intelligent Robots and Systems (IROS 2019), November 4 - 8, 2019, Macau, China}

\end{onecolumn}

\twocolumn

\maketitle

\begin{abstract}
Robotic pollinators not only can aid farmers by providing more cost effective and stable methods for pollinating plants but also benefit crop production in environments not suitable for bees such as greenhouses, growth chambers, and in outer space. Robotic pollination requires a high degree of precision and autonomy but few systems have addressed both of these aspects in practice. In this paper, a fully autonomous robot is presented, capable of precise pollination of individual small flowers. Experimental results show that the proposed system is able to achieve a 93.1\% detection accuracy and a 76.9\% `pollination' success rate tested with high-fidelity artificial flowers.
\end{abstract}



\section{Introduction}
Farmers are increasingly relying on technology to compensate for labor shortages and meet the growing demand for food. As a result, agricultural robotics are rapidly gaining interest in the research community and by the agriculture industry.
To meet the increasing demands of a growing human population, global food production must nearly double in the next few decades \cite{valin2014future}, which will require rethinking current agricultural practices.
The tasks involved in agriculture are often lengthy and repetitive making them well suited for robots.

In the past, automation in precision agriculture focused primarily on large-scale applications.
However, attention is also needed on precision tasks involving sensing and manipulation of individual plants for improved crop management and productivity.
The new generation of agriculture robots focus on plant parts (e.g. fruits, leaves, or flowers), which is necessary for automating tasks such as fruit and vegetable picking \cite{van2002autonomous, baeten2008autonomous, scarfe2009development, lehnert2017autonomous}, phenotyping \cite{mueller2017robotanist}, pollination \cite{ohi2018design, williamsautonomous, yuan2016autonomous}, and weed control \cite{slaughter2008autonomous, aastrand2002agricultural} to name a few. These applications require a high degree of precision and autonomy but few systems have addressed both of these two aspects in practice.

\begin{figure}[!t]
    \centering
    \includegraphics[width=\linewidth]{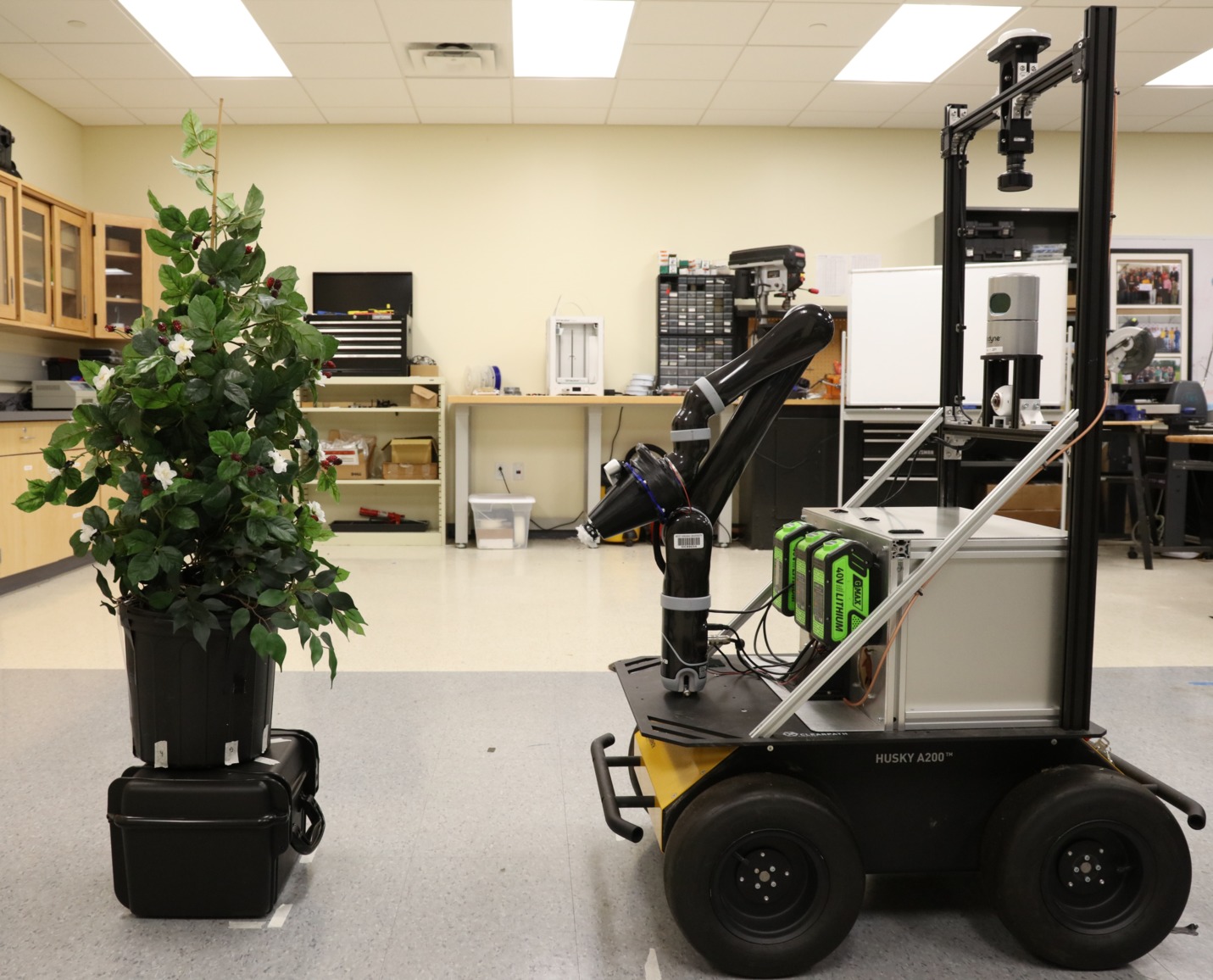}
    \caption{Experimental setup featuring the robotic arm with attached end-effector and depth-camera in front of an artificial bramble plant.}
    \label{fig:example_setup}
\end{figure}

One urgent challenge facing the agriculture industry is the decline of natural pollinators, which threatens the future of food production. As a result, many farmers cannot rely on wild pollinators and instead depend on services for renting bee colonies at a high cost.
Also, human introduced bee colonies may threaten wild pollinators due to competition for resources \cite{mallinger2017managed}.
While there is no major pollination crisis yet, there is evidence for localized limitation of crop yield as a result of inadequate pollination \cite{goulson2015bee}.
Robotic pollinators additionally benefit agriculture in environments not fit for natural pollinators such as greenhouses, growth chambers, and in outer space (e.g., in a Mars colony).
As a result, robotic pollinators can aid farmers by providing a more cost effective and stable method for pollinating plants as well as reduce the stress caused on rental bee colonies.

The idea of using robots to aid pollination has been considered for more than a decade \cite{binns2009robotic}; however, research in this area is quite limited beyond conceptual designs \cite{berman2011design, berman2011optimization, abutalipov2016flowering}, only a few systems have been demonstrated in practice \cite{shaneyfelt2013vision, gan2008stabilization,amador2017sticky}, and even fewer with autonomy \cite{williamsautonomous, yuan2016autonomous}. The systems developed in \cite{williamsautonomous, yuan2016autonomous} use sprayers for pollinating flowers of tomato and kiwifruits, respectively instead of physically touching each flower like bees would do.

In this work, we aim to fill the research gap by presenting a fully autonomous system capable of precise pollination of individual small flowers. The introduced system is developed as a subsystem for a ground vehicle such as the one presented in our previous work, BrambleBee \cite{ohi2018design}. BrambleBee is a fully autonomous robot developed for pollinating bramble plants (i.e., blackberry and raspberry) in a greenhouse environment. In our previous work, the pollination procedure was tested using AruCo markers \cite{garrido2014automatic} instead of actual flowers, which is addressed in this work. The system is tested with high-fidelity, artificial flowers and further experiments will be performed when real flowers bloom in the near future.

The remainder of the paper is structured as follows. The problem specifications are discussed in \autoref{sec:problem_description}. The general concept of the robot and software design are presented in \autoref{sec:system_overview}. A detailed description of the methods employed are presented for identifying flowers and estimating flower pose in \autoref{sec:perception}, mapping of the flowers and obstacles in \autoref{sec:mapping}, planning and control of the robotic arm in \autoref{sec:planning_control}, and the details behind the custom end-effector design in \autoref{sec:manipulation}. The experimental results are provided in \autoref{sec:experimental_results} and the conclusion and future work are discussed in \autoref{sec:conclusion}.

\section{Problem Description} \label{sec:problem_description}
A brief overview of the BrambleBee robot system and the assumptions are provided here to help the reader more easily understand the underlying system and the motivation of this work.

BrambleBee, as shown in \autoref{fig:example_setup}, is a ground vehicle designed for autonomously pollinating flowers in a greenhouse environment \cite{ohi2018design}. Built upon a ClearPath Robotics$^{\tiny{\textregistered}}$ Husky platform, the vehicle is equipped with a robotic arm (KINOVA$^{\tiny{\textregistered}}$ JACO 2) mounted to the front edge of the vehicle.
Attached to the robotic arm are two components: a custom designed end-effector utilized for pollinating flowers and a depth-camera (Intel$^{\tiny{\textregistered}}$ RealSense\textsuperscript{TM} D435) utilized for mapping the local environment (i.e., the workspace).

BrambleBee operates in a greenhouse environment with plants arranged in rows, so the robot is able to examine plants on each side.
Initially, BrambleBee explores the greenhouse to inspect the plants and construct a map of the environment. After a map is created, BrambleBee visits plants with flowers and executes the pollination procedure. Specifically, this paper presents the detailed procedure for robotic pollination, along with experimental results using high-fidelity, artificial flowers.

The case is considered where BrambleBee is parked in front of the plants. The goal of the proposed subsystem is to pollinate all flowers reachable by the end-effector attached to the robotic arm.

\begin{figure}[!t]
    \centering
    \vspace*{0.25cm}
    \includegraphics[width=\linewidth]{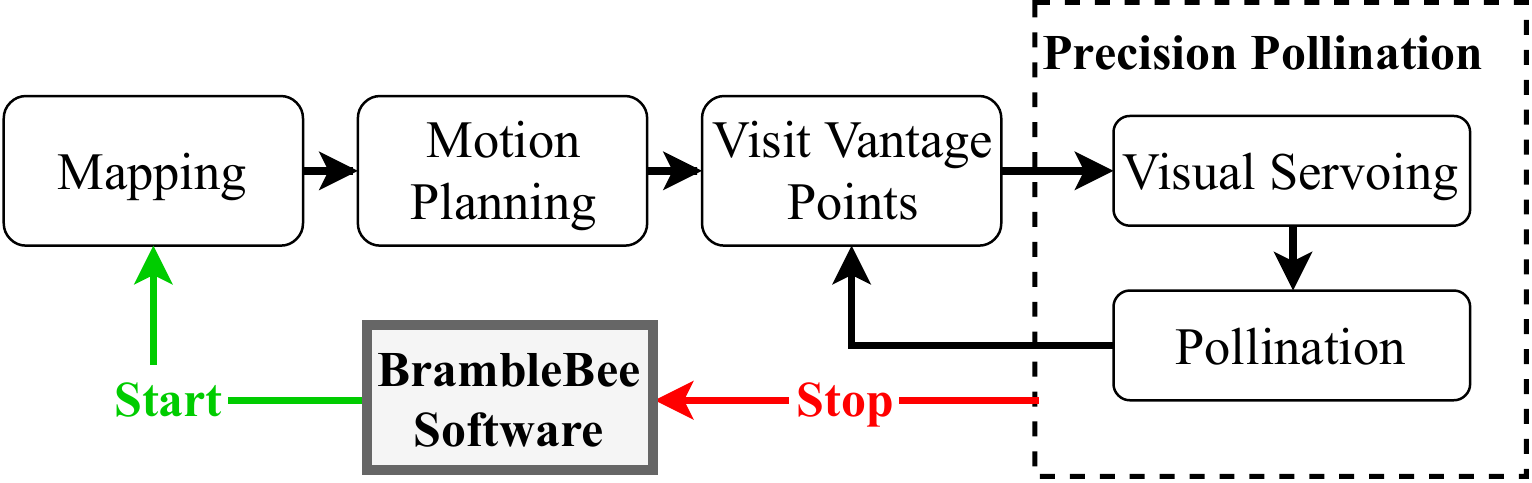}
    \caption{Diagram of the overall concept of operations developed for automating the system and integrating the separate software components. Activation of the manipulation system comes from BrambleBee's full system and signals the full system when pollination procedures are complete.}
    \label{fig:concept_operations}
\end{figure}

\section{System Overview}  \label{sec:system_overview}

As depicted in \autoref{fig:concept_operations}, the proposed system is activated by BrambleBee after the robot is positioned in front of the plants.
Once activated, the system starts by mapping the flowers and obstacles in the workspace. This is achieved by manuevering the end-effector through a set of poses that cover the workspace.
At each end-effector pose, image processing algorithms are applied to identify flowers and estimate the corresponding flower poses. Concurrently, the depth information is used to map the obstacles in the workspace (e.g., other plant parts or structures). The resulting obstacle map is used to avoid possible collisions that could damage the plant or robot.
After mapping the workspace, a trajectory is planned for the end-effector through a set of vantage points in front of each flower.

At each vantage point, the pose is refined before pollinating the target flower by collecting and fusing additional pose estimates using a factor graph based framework \cite{dellaert2017factor}.
Using the refined pose, the end-effector aligns itself to the flower and activates the visual servoing procedure to guide the end-effector towards the flower until contact is made.
Once reached, the precision pollination procedure is executed.
This operation actuates the end-effector to perform a motion that allows the pollen to be released from the anthers of the flower.
Note that bramble flowers can be pollinated using pollen from the same or other bramble flowers. This process is repeated until all flowers in the workspace are pollinated.

\begin{figure*}[!t]
    \centering
    \vspace*{0.25cm}
    \includegraphics[width=\linewidth]{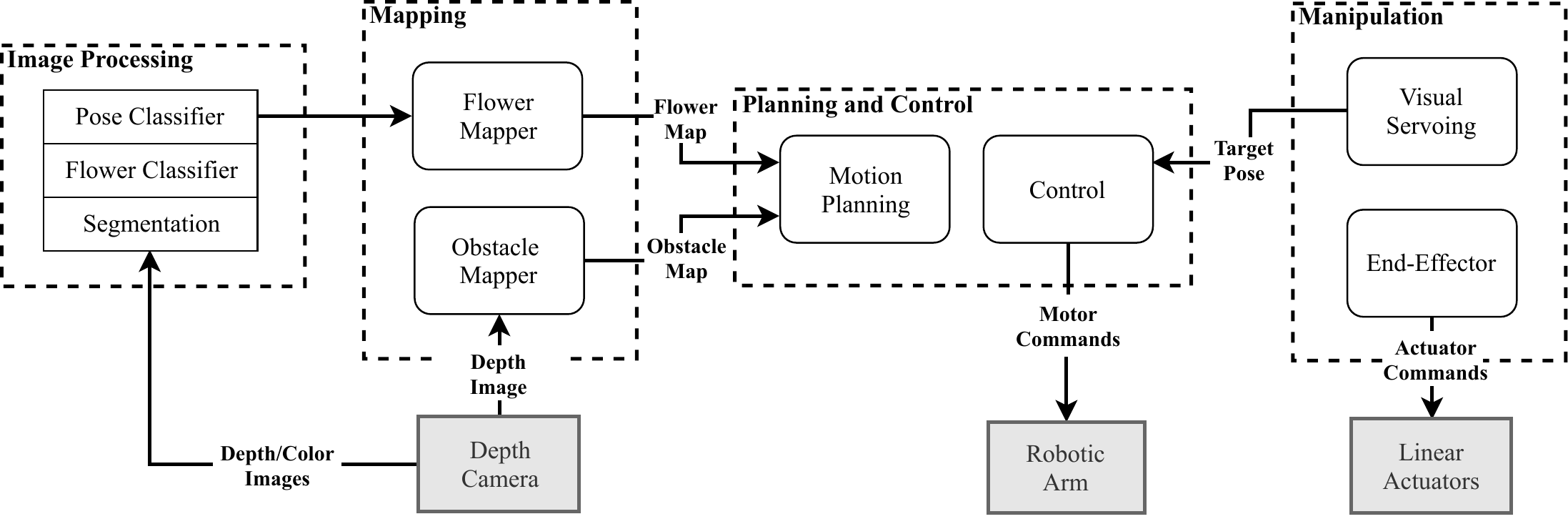}
    \caption{An overview of the software architecture developed for the system. The software is managed through a finite-state machine and separated in four components including image processing, mapping, manipulation, and planning and control.}
    \label{fig:software_architecture}
\end{figure*}

The primary software modules as well as their relationships are highlighted in \autoref{fig:software_architecture} and are managed through a finite-state machine. The software architecture of the system can be divided into four primary modules: 1) image processing, 2) mapping, 3) manipulation, and 4) planning and control. The image processing module is responsible for identifying flowers in the environment and estimating the corresponding flower poses.
The manipulation module is utilized for generating the precise movements of the end-effector required to pollinate each flower.
The planning and control module is responsible for motion planning and control of the robotic arm.
A detailed discussion of the employed algorithms are provided in the following sections.

\section{Image Processing} \label{sec:perception}
The robotic pollination system must accurately identify and estimate the pose of each flower. The proposed system achieves this through a two-stage framework consisting of a segmentation step followed by a classification step. The segmentation step extracts patches from the images acquired from the depth-camera based on color in order to reduce the search space for the classification algorithm. This step not only reduces the required computation of the entire pipeline but also improves the classification accuracy. The classification step is used to distinguish between flower and non-flower patches as well as estimate the pose of the identified flowers.

\subsection{Naive Bayes' Pixel-Level Segmentation} \label{sec:framework}
The segmentation step is used to classify each pixel based on color as belonging or not belonging to part of a flower.
A naive Bayes' classifier \cite{11,52} is chosen for this step for several reasons. First, naive Bayes' classification provides a direct prediction of the posterior probabilities of the class labels avoiding manual parametrization.
Second, naive Bayes' classification is robust to missing information, and as a result, the feature space is well represented by a modest number of diverse training images.
Therefore, a naive Bayes' classifier is applied to segment the image before applying the transfer learning based classifier discussed in the following sections.

In general, the naive Bayes' classifier is a family of conditional probability models based on applying Bayes' theorem with the assumption of conditional independence among features.
In this case, the pixel intensities are considered as features; therefore, after applying Bayes' theorem with the independence assumption, the joint model can be expressed as
\begin{equation}
	p(l|u) = \frac{p(l) p(u|l)}{p(u)}
\end{equation}
where $p(u|l)$ can equivalently be written as $p(r|l)p(g|l)p(b|l)$ where $l$ is the class label and $r$, $g$, and $b$ are the intensities of the red, green, and blue channels respectively. Therefore, $p(l|u) \propto p(l) p(u|l)$, and the classification rule for a given pixel is then given by
\begin{equation}
	\hat{l} = \argmax_{l \in L} p(l) p(u|l)
	\label{map}
\end{equation}
where $\hat{l}$ is the Maximum A Posteriori (MAP) estimate of the class label for a given pixel assuming conditional independence between pixel intensities. The priors $p(l)$ and the likelihoods $p(u|l)$ can be determined by calculating the relative frequency of the pixels in the training images.

To reduce the required computation per image, a lookup table $H$ is computed using all possible values for a pixel (e.g. 24 bits for most color images). The lookup table can then be accessed using the raw pixel values to efficiently segment the image. Therefore, $H$ using naive Bayes' classification is given by
\begin{equation}
	h_i = \argmax_{l \in L} p(l) p(q_i|l)
\end{equation}
for all $q_i \in \{0,1\}^b$ where $h_i \in H$ and $b$ is the number of bits for a single pixel. To prevent a bias towards the training images with higher resolution, the relative frequencies are normalized for each training image.

\subsection{Refinement of Segmentation using Convolutional Neural Networks (CNNs)} \label{sec:inception}
\begin{figure}[!t]
    \centering
    \vspace*{0.25cm}
    \includegraphics[width=\linewidth]{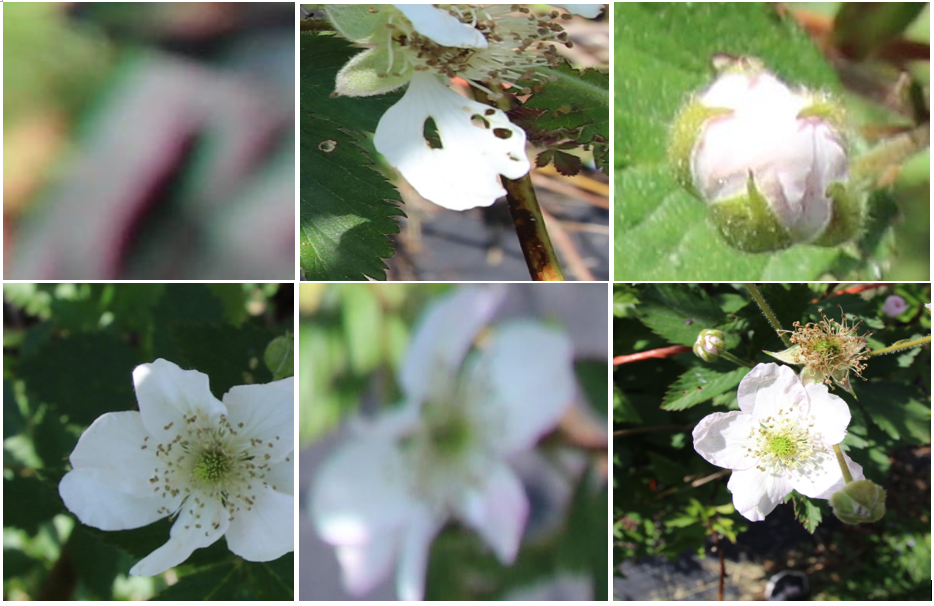}
    \caption{Examples of classification applied to image patches extracted from the segmentation algorithm. The patches in the top row are classified as non-flower with probabilities 99.8\%, 61.2\%, and 61.2\%, respectively. The patches in the bottom row are identified as flower with probabilities 91.1\%, 97\%, and 84.3\%, respectively.}
    \label{fig:example_classification}
\end{figure}

The segmentation step produces a set of patches for each image consisting of flowers and non-flowers. Thus, a method is proposed using machine learning to identify true (flowers) and false (non-flowers) positives extracted in the segmentation step. Inception-v3 \cite{53} is used for refining the segmentation. It computes the probability of each label
\(k \in \{1...K\}\):
\begin{equation}
	p(k|x) = \frac{\exp{(z_k)}}{\sum_{i=1}^{K} \exp{(z_i)}}
\end{equation}
where $x$ is a training example, $z_i$ is the logits or unnormalized log probability of each class \cite{53}, and $k$ is either flowers or non-flowers in this context. The loss function is defined as
\begin{equation}
	\ell=-\sum_{k=1}^{K} \log(p(k))q(k)
\end{equation}
where $q(k)$ is the ground-truth distribution. The above cross entropy loss function is differentiable with respect to the log probability $z_k$ which allows the use of gradient descent for training the neural networks. The gradient is bounded between -1 and 1 and has the following form:
\begin{equation}
	\frac{\partial \ell}{\partial z_k}=p(k)-q(k).
\end{equation}
In our approach, a transfer learning technique was adopted by taking advantage of the body of Inception-v3, which has rich features. The softmax layer was modified by retraining the network to perform binary classification. In order to train the network, the positive and negative patches were obtained by comparing initial segmentation results against manually labeled images. There are 13,395 positive and 15,066 negative patches extracted in total from the labeled images. The training took around 35 minutes using an Intel i7-4790k CPU and an NVIDIA Titan X GPU in Tensorflow. The results for a set of patches not included in the training data are presented in \autoref{fig:example_classification}.

\subsection{Pose of Flowers}
\begin{figure}[!t]
    \centering
    \vspace*{0.25cm}
    \begin{subfigure}[b]{0.3\linewidth}
        \includegraphics[width=\textwidth]{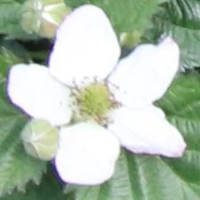}
        \caption{c1}
        \label{fig:gull}
    \end{subfigure}
    ~
    \begin{subfigure}[b]{0.3\linewidth}
        \includegraphics[width=\textwidth]{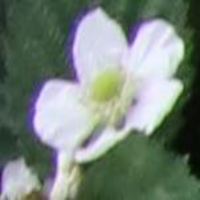}
        \caption{c2}
        \label{fig:tiger}
    \end{subfigure}
    ~
    \begin{subfigure}[b]{0.3\linewidth}
        \includegraphics[width=\textwidth]{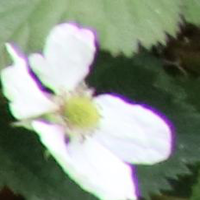}
        \caption{c3}
        \label{fig:mouse}
    \end{subfigure}
    \caption{Examples of orientation classes where the center of the flower is pointing at the center of the camera $c_1$, towards the left of the camera $c_2$ , and towards the right of the camera $c_3$.}\label{fig:poseExam}
    \label{fig:flower_pose_examples}
\end{figure}

For the end-effector to accurately reach the center of each flower, the pose is estimated for each identified flower. The position of each flower is extracted from the pixel coordinates and corresponding depth using back-projection given the intrinsic camera parameters. In contrast, the orientation is not observed directly; thus, a learning approach is implemented to approximate the orientation of each flower. In general, the center of a flower may point toward any arbitrary direction; however, the end-effector (discussed in \autoref{sec:manipulation}) is designed to allow for error in the flower orientation. Therefore, we simplify the flower orientation into three classes: the center points towards the center of the camera $c_1$, towards the left of the camera $c_2$, and towards the right of the camera $c_3$ as shown in \autoref{fig:flower_pose_examples}.

This allows us to formulate the problem of determining the orientation of each flower as a multi-class classification problem, which can be solved using CNNs. Similar to the method applied for refining the segmentation, the orientation is determined by training an Inception-v3 network with three classes. Our experiments with real flower patches show that the classifier could reach approximately 70\% precision and recall for orientation. A summary of the performance is given in \autoref{tab:classification_results}. In the future, additional data will be collected to improve the accuracy of both the flower and pose classifiers. To reduce estimation errors, multiple observations are fused, which improve the pose estimates of each flower. This is discussed more in the following sections.
\begin{table}[!t]
    \caption{Flower and Orientation Classification Results}
    \centering
    \begin{tabular}{p{0.9cm}p{0.7cm}p{1.1cm}p{0.9cm}p{1.2cm}p{0.9cm}}
        \toprule
         & Class & Training & Testing & Precision & Recall\\
         \midrule
          \textbf{Flower} & Pos & 13,395 & 2,102 & 78.6\% & 90\%   \\
                          & Neg & 15,066 & 2,124 & 88.5\% & 75.8\% \\
        \midrule
        \pbox{20cm}{\textbf{Orient-} \\ \textbf{ation}}     & C1 & 796 & 60 & 79.3\% & 83.3\%  \\
                  & C2 & 920 & 88 & 74.3\% & 59.1\%  \\
                                 & C3 & 771 & 72 & 59.5\% & 61.1\%  \\
        \bottomrule
    \end{tabular}
    \label{tab:classification_results}
\end{table}
%

\section{Mapping}  \label{sec:mapping}

\subsection{Obstacle Map}
The obstacle map is used for motion planning to avoid collisions with the plant or other objects in the environment. The map is represented as a 3D occupancy grid where each voxel represents the probability that the voxel is occupied by an object. In this work, we use the octree-based mapping framework \cite{hornung2013octomap} where the voxels are managed as a tree allowing for compact memory representation and multiple query resolutions.

\begin{figure}[!b]
    \centering
    \includegraphics[width=0.9\linewidth]{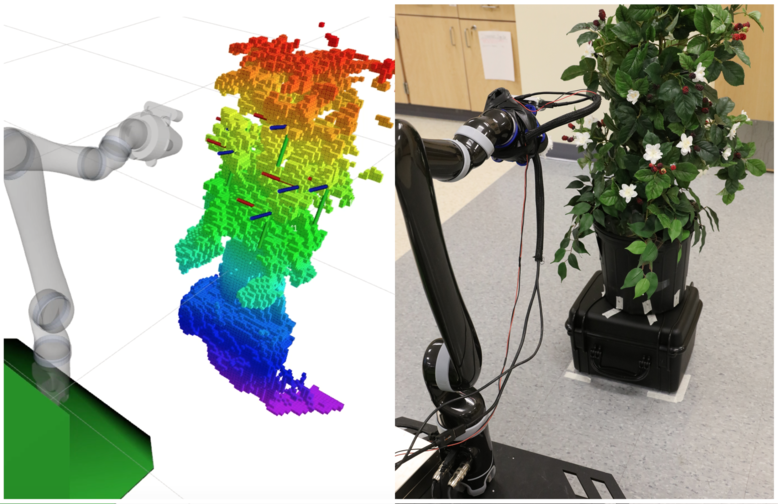}
    \caption{Example of an occupancy map estimated during the mapping procedure of a single plant with models for the robotic arm and base of BrambleBee. The left image illustrates the identified flowers in the occupancy map during the experiment displayed in the right image.}
    \label{fig:example_occupancy_map}
\end{figure}

To map the workspace, the mapping procedure is performed by moving the arm through a predefined set of poses such that the sensor (i.e., depth-camera) observations will cover the space reachable by the end-effector. As the arm moves through the set of poses, the obstacle map is continuously updated as measurements are acquired by the depth-camera mounted on the robotic arm.
An example of the obstacle map estimated using a set of 10 predefined poses is presented in \autoref{fig:example_occupancy_map}.

\subsection{Flower Map}
When performing robotic pollination, a flower map is maintained that contains the pose of each flower observed using the perception algorithms described in \autoref{sec:perception}.
Specifically, a factor graph representation is utilized -- as depicted in \eqref{fgCost} -- to partition the posterior distribution into three subsets: the prior information about the workspace $\psi_p$, the dynamic information about the workspace $\psi_d$, and the likelihood constrains about the workspace $\psi_l$.

\begin{equation}
\begin{split}
 X^{\text{MAP}} &= \argmax_x \ P(X|Z) \ \\ &= \ \argmax_x \ \lbrace \prod_{i=1}^{I} \psi_{p,i}  \prod_{j=1}^{J} \psi_{d,j}  \prod_{k=1}^{K} \psi_{l,k} \rbrace
 \label{fgCost}
\end{split}
\end{equation}

The optimization problem represented in \eqref{fgCost} can be further simplified when it is assumed that the system dynamics and the collected measurements are only corrupted by additive Gaussian noise. When this assumption holds, the optimization problems simplifies to a Non-Linear Least Squares (NLLS) problem, as presented in \eqref{fgNLLS},

\begin{align}
\begin{split}
 X^{\text{MAP}} = \argmin_{X}  \ \bigg[ \ &\sum_{i=1}^{I} \lvert \lvert x_o - x_i \rvert \rvert^{2}_{\Sigma} \ + \\ & \sum_{j=1}^{J} \lvert \lvert x_j - f_j(x_{j-1}) \rvert \rvert^{2}_{\Lambda} \ + \\ & \sum_{k=1}^{K} \lvert \lvert z_k - h_k(x_{k}) \rvert \rvert^{2}_{\Xi} \bigg],
 \label{fgNLLS}
 \end{split}
 \end{align}
where $x_o$ is the prior information about the workspace, $f_j$ incorporates the knowledge of workspace dynamics, and $h_k$ is the observation mapping function. Additionally, $\Sigma$ incorporates the uncertainty about the prior information, $\Lambda$ incorporates the uncertainty about the system dynamics, and $\Xi$ incorporates the uncertainty about the measurements, respectively.

To adopt the generic formulation presented in \eqref{fgNLLS}, to the problem of flower pose filter, we first assume a static motion model; however, additional information about the growth cycle could be incorporated later. Additionally, the set of observations $\{z\}_{k=1}^K$ are provided by the previously specified pose classifier. Utilizing the specified models and observations, the cost function provided in \eqref{fgNLLS} is optimized using the Levenberg-Marquardt algorithm \cite{more1978levenberg} to provide a filtered pose estimate of each flower in the workspace.

\begin{figure*}[!h]
    \centering
    \vspace*{0.25cm}
    \includegraphics[width=\linewidth]{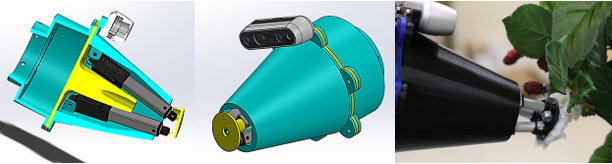}
    \caption{(Left) Section view of the end-effector showing two of the three linear servos inside the 3D printed housing mounted to the base of the end-effector. (Middle) Outer view of the end-effector showing the flexible plate connected to the linear actuators. (Right) Outer view of the end-effector performing the pollination procedure during experiments.}
    \label{fig:end_effector_design}
\end{figure*}

\section{Planning and Control}  \label{sec:planning_control}

\subsection{Motion Planning}
After estimating the flower and obstacle maps, a trajectory is planned for the end-effector to visit and pollinate each flower. The goal is to find a trajectory that minimizes the motion required to visit each flower under the kinematic constraints of the arm given the pose of each flower and the obstacle map. As described in \autoref{sec:system_overview}, a vantage point is defined in front of each flower to refine the pose of the flower before pollination. The set of vantage points are denoted $\boldsymbol{v} = \{v_1, v_2, \cdots, v_N \}$ where $v_i$ is the pose of the end-effector at the $i$th vantage point and $N$ is the number of flowers in the map. The path length (or cost) of the end-effector to travel between a pair of vantage points is given by
\begin{equation}
    c_{i,j} = \int_{}^{} \lvert \lvert \boldsymbol{\pi}_{i,j}(t) \rvert \rvert \,dt
    \label{eq:path_cost}
\end{equation}
where $\boldsymbol{\pi}_{i,j}(t)$ is a continuous-time function defining the trajectory of the end-effector between vantage points $v_i$ and $v_j$ determined by a point-to-point planner (e.g. planning the trajectory between a pair of vantage points). In this work, the Open Motion Planning Library (OMPL) \cite{sucan2012open} is utilized for point-to-point planning.

The problem of finding the shortest end-effector path through all vantage points is a form of the \ac{TSP} with the corresponding planning objective defined as
\begin{equation}
    J(o) = \sum_{i=1}^{\lvert o \rvert - 1} c_{o(i),o(i+1)}
    \label{eq:planning_objective_trajectory}
\end{equation}
where $o \in \mathcal{O}$ is some ordering of $\lvert o \rvert$ vantage points such that $\mathcal{O}$ is the set of all permutations of vantage points and $J(o)$ is the corresponding cost for visiting each vantage point for some ordering $o$. From the previous sections, we know the pose of each flower as well as the space occupied by obstacles. Using this information, each vantage point is set as a constant offset, so each flower is in view of the camera from the corresponding vantage point. Therefore, the following optimization problem is solved:
\begin{equation}
    \{o\} = \argmin_{o \in \mathcal{O}} J(o).
    \label{eq:planning_objective_trajectory}
\end{equation}
Several software packages were used in implementing the proposed methods. The inverse kinematics of the arm were solved using TRAC-IK \cite{beeson2015trac}. To check for collisions during planning, the \ac{FCL} \cite{pan2012fcl} is utilized, which incorporates the model of the arm and the generated obstacle map. These libraries were encapsulated in the software library MoveIt! \cite{chitta2012moveit} and were used in the software developed for motion planning.

\subsection{Visual Servoing} \label{sub:vs}

Once the end-effector is positioned in front of a flower, visual servoing is used to steer the end-effector towards the flower by controlling the trajectory in terms of desired end-effector positions. The procedure is comprised of mainly two steps: 1) The axis of the end-effector is aligned with the center of the flower by moving in the plane parallel to the face of the flower; 2) The end-effector moves along the axis orthogonal to the flower until making contact.
In order to execute this procedure, the velocities for each individual joint $\Dot{\mathbf{q}}$ are determined such that the end-effector reaches the desired pose. The joint velocities are computed from a vector of end-effector translational and angular velocities (i.e., $\mathbf{v}^*$ and $\mathbf{\omega}^*$, respectively) denoted by $\Dot{\mathbf{x}}^* = [ \mathbf{v}^* \, \mathbf{\omega}^*]^T$.
The relationship between $\Dot{\mathbf{q}}$ and $\Dot{\mathbf{x}}^*$ is defined by
\begin{equation} \label{eq:vs1}
    \Dot{\mathbf{x}}^* = \mathbf{J}(\mathbf{q}) \Dot{\mathbf{q}}
\end{equation}
where $\mathbf{J}(\mathbf{q})$ is the robot Jacobian, which was found using TRAC-IK \cite{beeson2015trac}.

To achieve parallel servoing, the distance is computed in the plane parallel to the face of the flower $\mathbf{d}_{\mathbin{\|}}^g$ that aligns the end-effector with the flower. This is used to set the direction of the end-effector velocity such that $\Dot{\mathbf{x}}^* = \alpha [ \mathbf{d}_{\mathbin{\|}}^g \, \, \mathbf{0} ] ^T$ where $\alpha$ is a scalar representing the velocity scale. Since the proper orientation is assumed at the start of visual servoing, the angular velocities are set to zero except in the case where $\mathbf{J}(\mathbf{q})$ is ill-conditioned, which is discussed later.
The individual joint velocities are determined using
\begin{equation} \label{eq:vs3}
    \Dot{\mathbf{q}} = \mathbf{J}(\mathbf{q})^{-1} \Dot{\mathbf{x}}^*.
\end{equation}

During visual servoing, the norm of the joint velocities is set to a constant value, which determines the value of $\alpha$. This is always done before applying $\Dot{\mathbf{q}}$ to the joints to ensure safe and consistent performance of the arm, although this causes some variance in the velocity. When $\lVert \mathbf{d}_{\mathbin{\|}}^f \rVert$ is close to zero, this indicates that the end-effector and the center of the flower are nearly collinear (i.e., the end-effector is pointing almost directly at the center of the flower). Then, the procedure transitions to orthogonal servoing, which moves the end-effector towards the flower along the line orthogonal to the face of the flower. The global distance $\mathbf{d}^g$ from the tip of the end-effector to the center of the flower is used to set the direction of the velocity by setting $\Dot{\mathbf{x}}^* = \alpha [ \mathbf{d}^g \, \, \mathbf{0} ]^T$ where \eqref{eq:vs3} is used to determine $\Dot{\mathbf{q}}$.

Occasionally, the arm reaches singularity conditions in which the end-effector cannot move in the desired translational direction while maintaining a fixed orientation. Therefore, a check is used to determine if $\mathbf{J}(\mathbf{q})$ is ill-conditioned. If this is the case, translation-only servoing is performed to bypass the singularity condition. The Jacobian is reduced $\mathbf{J}_R(\mathbf{q})$ to be equal to the first 3 rows of the original $\mathbf{J}(\mathbf{q})$, then $\Dot{\mathbf{q}}$ is calculated using the Moore-Penrose right pseudo-inverse \cite{albert1972regression}:
\begin{equation} \label{eq:vs5}
    \Dot{\mathbf{q}} = \mathbf{J}_R(\mathbf{q})^T (\mathbf{J}_R(\mathbf{q}) \mathbf{J}_R(\mathbf{q})^T)^{-1} \mathbf{v}^*.
\end{equation}
This solution minimizes the effort $\lVert \Dot{\mathbf{q}}\rVert^2$ while still satisfying $\mathbf{v}^* = \mathbf{J}_R(\mathbf{q}) \Dot{\mathbf{q}}$.

This process is incrementally repeated until contact is assumed to be made with the flower. Due to the current design of the end-effector, the depth-camera eventually loses sight of the flower while approaching it. Therefore, there is a short motion where the manipulator blindly operates using the most recent flower pose estimate. In the future, an endoscope camera will be centrally placed in the end-effector to allow for continuous tracking of the flower until it is reached.


\begin{table*}[!t]
\centering
\vspace*{0.25cm}
\caption{Experimental results where `\# Trials' is the number of trials ran for each scenario, `\# Reachable' is the number of reachable flowers that are $\leq$ 0.7 m away from the base of the manipulator, `\# Avg. Seen' is the average number of flowers seen in the workspace, `\% Touched' is the percentage of flowers where the end-effector touched the flower, `\% Pollinated' is the percentage of flowers pollinated where the end-effector touched the flower and its anthers, and `\% Missed' is the percentage of flowers where the end-effector was not able to touch the flower.}
    \begin{tabular}{p{2.8cm}p{1.4cm}p{1.4cm}p{1.4cm}p{1.4cm}p{1.4cm}p{1.4cm}p{1.4cm}p{1.4cm}}
        \toprule
        \textbf{Scenario}       &   1   &   2       &   3      &   4       &   5       &   6   &   7      &   8      \\
        \midrule
        \textbf{\# Trials}   &   5   &   5       &   6       &   6       &   5       &   7   &   7   &   6    \\
        \textbf{\# Reachable}   &   3   &   3       &   2       &   2       &   2       &   4   &   4   &   4    \\
        \textbf{\# Avg. Seen}   &   3   &   2.6     &   2.8     &   1.8     &   2       &   3.7 &   3.4 &   3.8    \\
        \textbf{\% Touched}     &   100\% &   100\% &   70.6\%  &   100\%  &   100\%    &   100\%&   62.5\%&   91.3\% \\
        \textbf{\% Pollinated}  &   80\% &   76.9\% &   52.9\%  &   81.8\%  &   90\%    &   92.3\%&  62.5\% &   73.9\% \\
        \textbf{\% Missed}     &   0\% &   0\%     &   29.4\%  &   0\%     &   0\%     &   0\% &   37.5\% &   8.7\% \\
        \hline
    \end{tabular}
\label{tab:experimental_results}
\end{table*}

\section{Manipulation} \label{sec:manipulation}

\subsection{Mechanical Design}
The design of the end-effector was inspired by a mixture of natural pollinators and human pollination methods.
The end-effector must be capable of reaching a desired pose with millimeter accuracy without damaging the plant or flowers.
Several key constraints were considered while designing the end-effector such as the range of actuation, size, and material.
Due to the size of the bramble flowers, the diameter of the tip of the end-effector is limited to no more than 4 cm.
The tip must also be flexible to enable an increased range of motion, which allows for the precise alignment of the end-effector to each flower.

To achieve this, we use three miniature linear servos (Actuonix L16-R) inside a 3D printed enclosure acting as a parallel robot.
A flexible plate is attached to the linear servos to allow for off-axis flexibility.
The material used for the plate is TPU-95, which is flexible and allows a wide range of motion. It is then coated in cotton padding for transferring pollen. In the future, alternative materials will be investigated for attachment to the end-effector.

\subsection{Inverse Kinematics}
Due to the flexible nature of the tip of the end-effector, a lookup table was employed for approximating the inverse kinematics to enable precise pose control.
To create the lookup table, the end-effector ran through all permutations of actuator commands, and the pose of the flexible plate (in the camera reference frame) and the pose of the joints (in the arms reference frame) was recorded for each permutation.
To record the end-effector pose, an AruCo marker was attached to the flexible plate on the end-effector and the Intel$^{\tiny{\textregistered}}$ RealSense\textsuperscript{TM} was used to extract the pose.
Using the recorded poses of the end-effector (in the camera reference frame) and the joints of the robotic arm (in the arms reference frame), the lookup table is generated using standard methods for hand-eye calibration \cite{tsai1989new} to estimate the transformation between the end-effector and robotic arm for each permutation of actuator commands.
The resulting transformations were stored in a lookup table that can be queried to find the actuator commands closest to a desired end-effector pose.


\section{Experimental Results} \label{sec:experimental_results}
We performed a series of experiments to evaluate the performance of the described system. Since these experiments were conducted during the winter months when bramble flowers were not in bloom, an artificial plant that resembles a real bramble bush, with high-fidelity, artificial bramble flowers, was used instead. The artificial plant was divided in 8 separate sections (or scenarios), where each section contained a varying number of isolated flowers.
Earlier, the experimental setup was illustrated in \autoref{fig:example_setup}.
For each scenario, at least 5 trials were performed, giving a total of 47 experiments. The results shown in \autoref{tab:experimental_results} summarize respectively the number of trials for each scenario, the number of reachable flowers (i.e., flowers that are $\leq$ 0.7 m away from the base of the manipulator), the average number of flowers seen in the workspace, the percentage of flowers touched, the percentage of flowers `pollinated' for each scenario (i.e., the end-effector touched the flower and its anthers after extending the linear actuators), as well as the percentage of flowers that were missed (i.e., not touch the flower after extending the linear actuators).

Out of the 144 total flowers in all trials, 134 of the flowers were accurately identified by the image processing algorithms, with only two false positives, yielding a 93.1\% detection accuracy. Our pollination success rate is 76.9\%. In the failed attempts, most flowers were either facing away from BrambleBee, in difficult to reach areas, or occluded by the plant leaves. During the failure cases, the tip of the end-effector would miss the center of the flower by no more than 2 cm. The main causes of these errors are: 1) errors in the estimated orientation of individual flowers and 2) the `blind driving' while approaching a flower since the depth-camera loses sight of the flower. Thus, improving the algorithms for estimating the pose of flowers will be a focus on future research. Also, as stated in \autoref{sub:vs}, errors due to `blind driving' towards a flower will be mitigated by utilizing an endoscope camera that will be centrally placed in the end-effector. This will allow for continuous flower tracking until contact with the flower.

\section{Conclusion and Future Work} \label{sec:conclusion}
This paper presented a fully autonomous system with precision pollination of small flowers. The proposed pollination system was developed as a subsystem for the autonomous ground vehicle BrambleBee.
Technologies in perception, planning and control, and autonomy were integrated to enable precise interactions with flowers. The proposed system has the potential to be leveraged for meticulous tasks such as harvesting and monitoring of crops.
The experiments show the robot is capable of operating with high precision and is able to achieve a 93.1\% detection accuracy and a 76.9\% pollination success rate on average. The capabilities of the developed system are demonstrated in this video: \texttt{https://youtu.be/ZbgtP9CHycA}. To our knowledge, this system is the first to demonstrate both precision and autonomy for pollinating small flowers.

A brief summary of the future work discussed throughout the paper is provided here. Currently, our pollination system works well for sparsely populated artificial flowers; however, the system will be verified in the near future through experiments on real plants once flowers are blooming. The primary failure mode of the system was missing contact with the anthers of a flower. This is due to errors in the pose estimates of each flower (particularly the orientation). Thus, further work is needed on estimating the pose of flowers, which would significantly increase the accuracy of the proposed system.


\bibliographystyle{IEEEtran}
\bibliography{main}

\end{document}